\documentclass[10pt,twocolumn,letterpaper]{article}
\usepackage{cvpr}              % To produce the CAMERA-READY version
\usepackage{multirow}
\usepackage[dvipsnames]{xcolor}

\definecolor{cvprblue}{rgb}{0.21,0.49,0.74}
\usepackage[pagebackref,breaklinks,colorlinks,citecolor=cvprblue]{hyperref}

\def\paperID{7403} % *** Enter the Paper ID here
\def\confName{CVPR}
\def\confYear{2024}

\title{Spatial Cascaded Clustering and Weighted Memory for 

Unsupervised Person Re-identification}

\author{Jiahao Hong
\qquad
Jialong Zuo
\qquad
Chuchu Han
\qquad
Ruochen Zheng
\qquad
Ming Tain\\
\qquad
Changxin Gao
\qquad
Nong Sang\\
{\tt\small \{hongjiahao, jlongzuo, cch,  cgao, zhengruochen, tainming, nsang\}@hust.edu.cn}
}

\begin{document}
\maketitle
\begin{abstract}
Recent unsupervised person re-identification (re-ID) methods achieve high performance by leveraging fine-grained local context. These methods are referred to as part-based methods. However, most part-based methods obtain local contexts through horizontal division, which suffer from misalignment due to various human poses. Additionally, the misalignment of semantic information in part features restricts the use of metric learning, thus affecting the effectiveness of part-based methods. The two issues mentioned above result in the under-utilization of part features in part-based methods. We introduce the Spatial Cascaded Clustering and Weighted Memory (SCWM) method to address these challenges. SCWM aims to parse and align more accurate local contexts for different human body parts while allowing the memory module to balance hard example mining and noise suppression. Specifically, we first analyze the foreground omissions and spatial confusions issues in the previous method. Then, we propose foreground and space corrections to enhance the completeness and reasonableness of the human parsing results. Next, we introduce a weighted memory and utilize two weighting strategies. These strategies address hard sample mining for global features and enhance noise resistance for part features, which enables better utilization of both global and part features. Extensive experiments on Market-1501 and MSMT17 validate the proposed method's effectiveness over many state-of-the-art methods.
\end{abstract}    
\section{Introduction}
\label{sec:intro}

Most recent unsupervised person re-identification methods  \cite{fan2018unsupervised,lin2019bottom,chen2021ice,wang2021camera,xuan2021intra,zhang2021refining,cho2022part,zhang2022implicit} have adopted an alternating two-stage training framework: 1) clustering samples in feature space to provide pseudo labels for training. 2) training the network under the supervision of pseudo labels generated by clustering. However, these methods are still susceptible to the limitations of clustering algorithms, as imperfect clustering methods may introduce noise.

\begin{figure}[t]
  \centering
  \begin{subfigure}[t]{0.31\linewidth}
    \centering
    \includegraphics[width=0.5\linewidth]{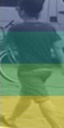}
    \caption{\centering{Horizontal Split}}
    \label{fig:short-horizeontal}
  \end{subfigure}
  \hfill
  \begin{subfigure}[t]{0.31\linewidth}
    \centering
    \includegraphics[width=0.5\linewidth]{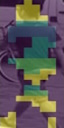}
    \caption{\centering{Feature Space Split}}
    \label{fig:short-feature}
  \end{subfigure}
  \hfill
  \begin{subfigure}[t]{0.31\linewidth}
    \centering
    \includegraphics[width=0.5\linewidth]{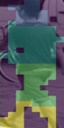}
    \caption{\centering{Ours}}
    \label{fig:short-parsing}
  \end{subfigure}
  \\
  \begin{subfigure}[t]{1\linewidth}
    \centering
    \includegraphics[width=1\linewidth]{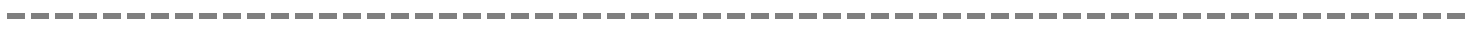}
  \end{subfigure}
  \\
  \begin{subfigure}{0.31\linewidth}
    \centering
    \includegraphics[width=1\linewidth]{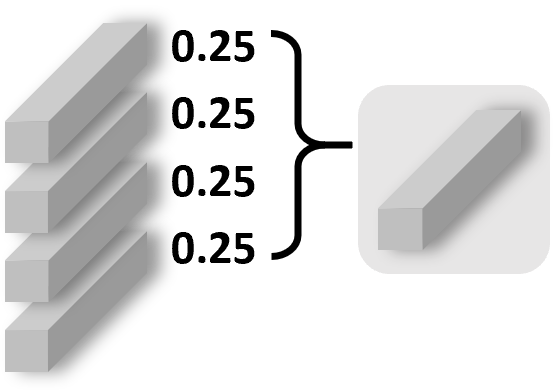}
    \caption{\centering{Average Update}}
    \label{fig:short-average}
  \end{subfigure}
  \hfill
  \begin{subfigure}{0.31\linewidth}
    \centering
    \includegraphics[width=1\linewidth]{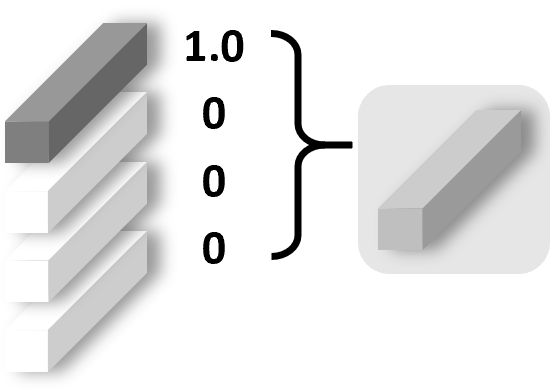}
    \caption{\centering{Hardest Update}}
    \label{fig:short-hardest}
  \end{subfigure}
  \hfill
  \begin{subfigure}{0.31\linewidth}
    \centering
    \includegraphics[width=1\linewidth]{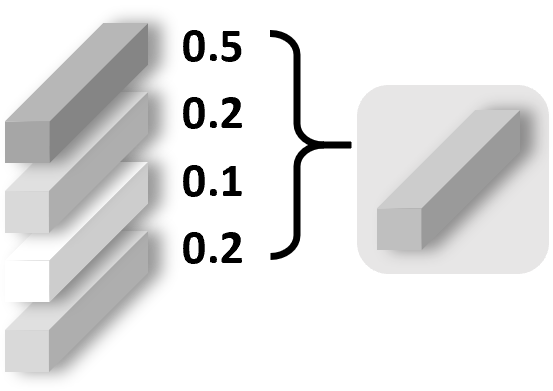}
    \caption{\centering{Ours}}
    \label{fig:short-weighted}
  \end{subfigure}
  \caption{Method differences illustration. The first row shows the difference in fine-grained split between existing methods and ours. Our method incorporates spatial information to correct the feature space, making the clustering results more coherent. The second row shows the difference in memory update strategy. Our method balances hard sample mining and noise resistance in the memory module through weighting.}
  \vspace{-3mm}
  \label{fig:short}
\end{figure}

The approach to addressing this issue categorizes existing methods into two types: one that utilizes only global information \cite{ge2020self,zhang2021refining,wang2020unsupervised,zheng2021online,ge2020mutual,zhang2022implicit} and the other that simultaneously uses global and local information \cite{cho2022part}. These two types of methods enhance the reliability and stability of clustering by leveraging different sources of information. The singularity of global features limits the first type of method, so we focus on the second type of method, which can leverage more information, namely part-based methods. Although part-based methods have achieved good results thanks to rich fine-grained information, their local partitioning methods are still too coarse \cite{cho2022part}. Furthermore, due to the differences in network structures and task settings, the direct application of previous human parsing methods may be limited \cite{zhu2020identity}, necessitating further corrections and improvements. 

Additionally, due to the absence of semantic alignment in local partitioning in previous methods, metric learning cannot be effectively applied to fine-grained local features \cite{li2017learning, cho2022part}, leading to limited network performance. Furthermore, previous research on metric learning has primarily focused on global features and has not considered the subtle differences between part and global features. Therefore, directly applying the previous memory module to local features has limited effectiveness.

In response to the problems above, we propose the Spatial Cascaded Clustering and Weighted Memory (SCWM) method. SCWM can automatically parse and align human body parts, and enable a more comprehensive utilization of global and local information. Building upon the existing part-based approach, our SCWM consists of a spatial cascaded clustering (SCC) and a weighted memory (WM).

Specifically, building upon previous work \cite{wu2021discover,zhu2020identity,chen2023beyond}, we utilize cascaded clustering as the pseudo parsing mask generator but enhance it through foreground and space corrections. In the first step, previous methods distinguish foreground and background by the strength of the feature map response. Still, they have ignored the neural network's tendency to prioritize the most salient regions, which can lead to regular foreground being mistaken for background. To address this issue, we cluster the parts with significantly high responses in the foreground separately. This helps avoid excessively high responses raising the average response of the foreground cluster, which might result in foreground with smaller responses being mistaken as background. In the second step, previous methods only cluster feature vectors in feature space, disregarding their positional relationships in spatial space. As a result, previous methods may incorrectly cluster similar features from different parts. Therefore, we opt for Agglomerative Clustering and utilize a precomputed distance matrix to incorporate spatial information. Given that a person's features in space are often not significantly distant, we assign an infinite distance between vectors if their spatial distance exceeds the threshold after calculating the distances between vectors in feature space. This spatial constraint significantly enhances the reliability of our clustering.

After aligning part features, we can employ metric learning to part features. However, initially designed for global features, the previous memory module \cite{ge2020self,dai2022cluster} shows limited performance when directly applied to part features. Therefore, we propose a weighted memory that simultaneously adapts to global and part features. Its flexible weighted update allows it to balance noise suppression and hard example mining. Given the differences between global and local features, we proposed two weighting strategies. For part features, due to the need for noise suppression, harder samples are considered to contain more noise. Therefore, the weights for hard samples should be smaller than simple samples. Conversely, for global features, the weighting for harder samples should be higher to construct samples that are more conducive to network learning. Additionally, previous methods only considered individual part spaces separately, neglecting the relationships between different part spaces. To address this limitation, we introduced a part separation loss, which aims to encourage part features to emphasize their unique information and enhance the overall diversity of part features.

Our contributions can be summarized as follows:
\begin{itemize}
    \item We improve the existing cascaded clustering by foreground correction and space correction.
\end{itemize}
\begin{itemize}
    \item We introduce a weighted memory and propose two weighting strategies to balance learning hard samples and reducing noise.
\end{itemize}
\begin{itemize}
    \item Extensive experimental results with superior performance against the state-of-the-art methods demonstrate the effectiveness of the proposed method.
\end{itemize}

\section{Related Work}
\label{sec:related work}

\textbf{Part-based person re-ID.} Part-based approaches for person re-ID leverage fine-grained information on human body parts, such as features from rigid stripe  \cite{sun2018beyond,zheng2019pyramidal}, auto-localization \cite{li2017learning,yao2019deep}, attention \cite{chen2019mixed,li2018harmonious,si2018dual,yang2019towards}, and extra semantics \cite{guo2019beyond,liu2018pose,miao2019pose,zheng2019camera,luo2023frequency,sarfraz2018a,song2018mask}. While part-based methods have demonstrated significant performance, the coarse part segmentation methods have limited these attempts to utilize part features. In visible-infrared re-ID, MPANet \cite{wu2021discover} proposed using lightweight convolutional layers to divide the feature map into different parts. However, the division results were not ideal due to the lack of clear semantic guidance. ISP \cite{zhu2020identity} introduced a self-learning human parsing method for the first time in supervised re-ID, which is used for pixel-level semantic part feature extraction. Recent work has also continued this line of thinking. SOLIDER \cite{chen2023beyond} employs semantic segmentation-generated masks of human body parts for self-supervised re-ID to perform image masking operations. What mainly sets our method apart from these is that when parsing human body parts, we consider the similarity between pixels in the feature space and the inherent continuity and separability of a person's body parts in spatial space.

\begin{figure*}[t]
  \centering
  \includegraphics[width=0.87\linewidth]{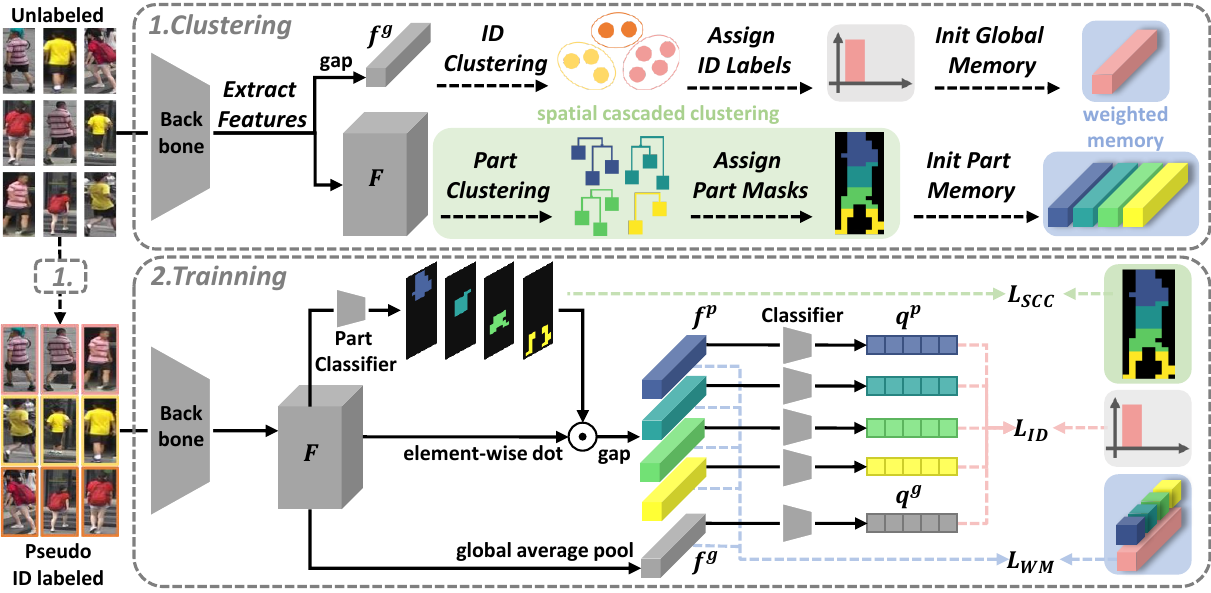}
  \caption{Framework of the proposed Spatial Cascaded Clustering and Weighted Memory (SCWM) method. The areas and lines annotated in light green represent the spatial cascaded clustering, while the areas and lines annotated in light blue represent the weighted memory.}
  \label{fig:overall}

\end{figure*}

\noindent\textbf{Memory module.} By establishing a memory module \cite{caron2020unsupervised,chen2020simple,he2020momentum,misra2020self,tian2020contrastive,wu2018unsupervised,ye2019unsupervised}, contrastive learning \cite{hadsell2006dimensionality} provides an efficient metric learning strategy in which the stored features can significantly reduce the difficulty of obtaining positive and negative sample pairs. Moco \cite{he2020momentum} has established a memory module for online updates of all sample features, significantly expanding the search scope for negative samples and overcoming the previous limitation of finding negative samples within a single batch. SpCL \cite{ge2020self} builds a memory module for both clustered samples and unclustered outliers. MCRN \cite{wu2022multi} proposes a memory module that can store multiple centers for a cluster, modeling the overall features of a cluster by considering information from multiple centers. CCL \cite{dai2022cluster} proposed a hardest sample updating strategy for the memory module. However, the method described above only considered global features, and the update strategy of the memory module was relatively simplistic, lacking the ability to adapt to current local features. Therefore, we proposed a weighted memory that can balance hard example mining and noise suppression through flexible weighted updates.

\section{Method}
\label{sec:method}

In this section, we begin with introducing our overall framework in \cref{method:overall}. Then, we discuss the spatial cascaded clustering in detail in \cref{method:parsing}. Next, we introduce the weighted memory and the two weighting strategies in \cref{method:memory}. Finally, we provide a brief overview of the adopted classification constraints and the overall object function of the model in \cref{method:classification}.

%-------------------------------------------------------------------------
\subsection{Overall Framework}
\label{method:overall}

We build our framework based on the current mainstream clustering-based method. All the following discussions will revolve around the two steps of the clustering-based method. The overall illustration is shown in \cref{fig:overall}.

\textbf{Clustering stage.} Firstly, we have an unlabeled dataset $\{x_i\}_{i=1}^N$ and need to extract their features. Here we freeze the parameter of the backbone network $f_\theta$ and use it to extract each sample's feature map $F$, global feature $f^g$, and part features $\{f^{p_k}\}_{k=1}^l$. $F$ is the output result of the last layer of the backbone. $f^g$ is the global average pooling result of $F$. For part features $f^{p_k}$,  we first use the backbone's part classifier $G(\cdot)$ to predict every part of the feature map. Then, we multiply these predicted part masks with the feature map and perform global average pooling to obtain part features of each part. The formulas are shown below:
\begin{equation}
    \begin{aligned}
        F & = f_\theta(x),\\
        f^g & = GAP(F),\\
        f^{p_k} & = GAP(G(F)_k \cdot F), k=1,2,...,l,
    \end{aligned}
\end{equation}
where $l$ is the number of parts in which the background is considered as one part, and $G(\cdot)$ consists of a $3\times 3$ conv layer followed by a softmax function. Secondly, we generate the required pseudo ID labels and pseudo parsing masks. We use DBSCAN \cite{ester1996density} clustering for all samples to assign a pseudo ID label to each cluster, denoted as $\{y_i\}_{i=1}^N$. For each sample, we apply cascaded clustering with foreground and space corrections to each feature map to obtain pseudo parsing masks, denoted as $M=[M_1,M_2,...,M_l]$. Finally, according to the pseudo ID labels, we individually initialize global and part memories with average cluster centroid. 

\textbf{Training stage.} We use the pseudo labels provided by the clustering stage to train the network during the training stage. Firstly, we extract the features for each sample following the same steps as in the clustering stage. Among them, we denote the predicted part masks generated by the part classifier as $P=[P_1,P_2,...P_l]$. Secondly, for the global feature and each part feature, we use separate classifiers to predict their categories vectors $q^g$ and $\{q^{p_k}\}_{k=1}^l$. Finally, we conduct parsing loss $L_{SCC}$, metric learning loss $L_{WM}$, and classification loss $L_{ID}$ on network $f_\theta$.

%-------------------------------------------------------------------------
\subsection{Spatial Cascaded Clustering}
\label{method:parsing}

Current unsupervised part-based person re-ID methods are constrained by misaligned part partitioning \cite{cho2022part}. Therefore, we introduce self-learning human parsing with spatial cascaded clustering to enhance the accuracy of part features. The difference is that, as previous methods applied directly to unsupervised tasks resulted in foreground omissions and spatial confusions, we specifically proposed foreground and space correction to address these two issues.

In the first step of cascaded clustering, upon observation, the feature map $F$ responses are consistently more significant for the foreground than the background \cite{mauthner2015encoding,zhu2020identity}. More importantly, some regions often have more salient responses within the foreground. However, due to the enormous responses of these salient regions, there is a risk of misclassifying foreground regions with smaller responses as background. Therefore, to mitigate the impact of salient regions on foreground judgment, we treat these regions as a distinct class during clustering. Hence, we adopt a three-class K-Means to divide the feature map $F$ into three parts: salient foreground $\{F_{sf}\}$, regular foreground $\{F_{rf}\}$, and background $\{F_b\}$, based on the magnitude of the feature vectors.

In the second step of cascaded clustering, we employ Agglomerative Clustering with a precomputed distance matrix adjusted by spatial information. This approach addresses the limitations of existing methods that do not utilize spatial information. To explore the distance relationships between the vectors at each position in the feature map and all other vectors, we denote the $c$-dimension vector at each position in the feature map as $v$. First, for all vectors in foreground $\{F_{sf}\} \cup \{F_{rf}\}$, we compute their distance matrix in spatial space and feature space, denoted as $D_s$ and $D_f$. Then, we set the distance in the feature space to infinity for vector pairs that are too far apart in spatial space, preventing confusion about a person's semantic features. Finally, we cluster $\{v| v\in\{F_{sf}\} \cup \{F_{rf}\}\}$ into $l-1$ parts $[M_1,M_2,...M_{l-1}]$ based on the corrected distance matrix $D$:
\begin{equation}
    D=\left\{
    \begin{array}{lr} 
      D_{f}(v_A,v_B), & if \quad D_{s}(v_A,v_B) < \eta,   \\
      \infty,         & if \quad D_{s}(v_A,v_B) \geq \eta,
    \end{array}\right.
\end{equation}
where $v_A$ and $v_B$ represent two vectors in $\{F_{sf}\} \cup \{F_{rf}\}$, $\eta$ is the threshold that determines distances that are considered too far apart in spatial space, $l$ is the total number of parts. The indices of the part clusters are arranged in descending order of the heights of the cluster centers. The background part $M_l$ is $\{F_b\}$.  To prevent abrupt changes in pseudo labels caused by variations in clustering between epochs, we employ the idea of momentum updating to smooth the pseudo parsing masks:
\begin{equation}
    \widetilde{M}\leftarrow\gamma\widetilde{M}+(1-\gamma)M,
\end{equation}
where $\gamma$ is the momentum factor, $\widetilde{M}$ is the smoothed pseudo parsing masks.

Next, in the training stage, we consider the pseudo parsing masks as supervision to guide the learning of the part prediction:
\begin{equation}
    L_{parsing}=-\widetilde{M}log(P).
\end{equation}

Furthermore, to ensure that different part masks can focus on distinct information, we use diversity loss to prevent the degradation of part masks:
\begin{equation}
    L_{diversity}=\frac{2}{l(l-1)} \sum_{i=1}^{l-1} \sum_{j=i+1}^{l} \sum_{(x,y)}(P_i \cdot P_j),
\end{equation}
where $P_i$ and $P_j$ mean the $i$-th and $j$-th predicted part mask, $(x,y)$ means the point in mask, $\cdot$ is element-wise multiplication. Then, the loss of spatial cascaded clustering is given by:
\begin{equation}
    L_{SCC}=L_{parsing}+L_{diversity}.
\end{equation}

%-------------------------------------------------------------------------
\subsection{Weighted Memory}
\label{method:memory}

Previous part-based methods do not apply metric learning to part features because the part features are not aligned \cite{sun2018beyond}. Now, after self-learning human parsing, the body parts of different individuals are semantically aligned, and we can employ metric learning for part features. Recent metric learning approaches \cite{ge2020self,dai2022cluster,zhang2022implicit} often employ a memory module to expand the search range for positive and negative samples in contrastive loss. However, these methods do not utilize part features, and their memory modules designed for global features have limited effectiveness when directly applied to part features. Therefore, we propose the weighted memory and two weighting strategies after exploring the differences between global and part features.

Specifically, in the clustering stage, we utilize K-nearest neighbors (kNN) similarity to measure the difficulty level of a specific feature within a sample. For the part feature of a sample, its difficulty level is determined by the Intersection over Union (IoU) between its K-nearest neighbors and the K-nearest neighbors of the sample's global features. As for global feature, its difficulty level is the average of all the difficulty levels of its sample's part features. We denote $\alpha$ as the difficulty level:
\begin{equation}
    \begin{aligned}
        \alpha^{p_k} = & \frac{kNN(f^g)\cap kNN(f^{p_k})}{kNN(f^g)\cup kNN(f^{p_k})}, \\
        \alpha^g = & \frac{1}{l} \sum_{k=1}^l \alpha^{p_k}.
    \end{aligned}
\end{equation}

Intuitively, a large $\alpha^{p_k}$ implies that the part feature is similar to the global feature, indicating that the information contained in the part feature is reliable and its difficulty level should be low. For a small $\alpha^g$, it indicates that the global feature does not fit well with the surrounding samples, suggesting that it contains feature information not yet accounted for in the current cluster. Therefore, its difficulty level should be high. In summary, the difficulty coefficient of global features should have an opposite trend to the $\alpha^g$, while the difficulty coefficient of part features should have the same trend as the $\alpha^{p_k}$. So when updating memory, we assign $1-\alpha^g$ for each global feature. In contrast, we assign $\alpha^{p_k}$ for each part feature. We use $l1$-normalization in a batch to obtain the final sample weights, ensuring the stable update of class centroids:
\begin{equation}
    \omega^g=\frac{1-\alpha^g}{\sum 1-\alpha^g_i}, \quad\omega^{p_k}=\frac{\alpha^{p_k}}{\sum \alpha^{p_k}_i},
\end{equation}
where $\alpha$ represents an individual sample's difficult level, while $\sum \alpha_i$ is the sum within a batch. Then, the momentum update formula for the weighted memory can be summarized in the following form:
\begin{equation}
    \begin{aligned}
        \forall f^g \in C_j^g, c_j^g & \leftarrow m c_j^g + (1-m)\omega^g f^g,\\
        \forall f^{p_k} \in C_j^{p_k}, c_j^{p_k} & \leftarrow m c_j^{p_k} + (1-m)\omega^{p_k} f^{p_k},
    \end{aligned}
\end{equation}
where $C_j^g$ and $C_j^{p_k}$ respectively stand for the set of global and $k$-th part features extracted from images of $j$-th cluster, $c_j^g$ and $c_j^{p_k}$ are the cluster centroids of global and $k$-th part features from $j$-th cluster, $m$ is the momentum factor.

Next, in the training stage, we modified the ClusterNCE loss \cite{dai2022cluster} into a weighted form:
\begin{equation}
    \begin{aligned}
        L_{wNCE}= - & log\frac{\omega^g \cdot exp(f^g\cdot c^g_+ / \tau)}{\sum_{j=1}^{N_C}{exp(f^g\cdot c^g_j / \tau)}} \\
    + \frac{1}{l}\sum_{k=1}^l - & log\frac{\omega^{p_k} \cdot exp(f^{p_k}\cdot c^{p_k}_+ / \tau)}{\sum_{j=1}^{N_C}{exp(f^{p_k}\cdot c^{p_k}_j / \tau)}},
    \end{aligned}
\end{equation}
where $\tau$ is temperature factor, $N_C$ is the number of clusters, $c_+$ is the centroid of the cluster to which $f$ belongs and $c_j$ is centroid of other cluster.

Moreover, given that different part features of the same individual should not overlap, we design a part separation loss in feature space:
\begin{equation}
    L_{sep}=-\frac{1}{l} \sum_{k=1}^l log\frac{exp(f^{p_k}\cdot c^{p_k}_+ / \tau)}{\sum_{j=1}^l{exp(f^{p_k}\cdot c^{p_j}_+ / \tau)}},
\end{equation}
where $c^{p_j}_+$ represents the cluster centroid of the $j$-th part feature of the samples in the cluster to which $f$ belongs.

Then, the loss of the weighted memory is given by:
\begin{equation}
    L_{WM}=L_{wNCE} + L_{sep}.
\end{equation}

%-------------------------------------------------------------------------
\subsection{Part-based Classification}
\label{method:classification}

Part-based classification methods have undergone significant development \cite{cho2022part, sun2018beyond, wang2018learning}. We follow the recent PPLR \cite{cho2022part} methods to design the classification constraints but explain them from a new perspective.

For a part feature, the lower its kNN similarity with the global feature, the more likely the part feature is noise. Therefore, its classification should not belong to any existing class. In formulaic terms, this can be seen as a form of weighted label smoothing, with the weighting factor being the sample difficulty coefficient:
\begin{equation}
    y^{p_k}_i=\alpha^{p_k}y_i + (1-\alpha^{p_k})uni,
\end{equation}
where $uni$ is a uniform vector representing a class label indicating that the feature does not belong to any class.

A form of knowledge distillation is needed for a global feature to incorporate the information from part features. This is reflected in the class labels by proportionally weighting the outputs of the part feature classifier to the global feature label:
\begin{equation}
    y^g_i=\beta y_i + (1-\beta)\widetilde{\omega}^{p_k}_i q^{p_k}_i,
\end{equation}
where $\beta$ is a hyper-parameter that controls the degree of knowledge distillation, $\widetilde{\omega}^{p_k}=\frac{exp(\alpha^{p_k})}{\sum_k exp(\alpha^{p_k})}$ is the weight that determines the proportion of each part.

Then, the classification loss can be summarized as follows:
\begin{equation}
    \begin{aligned}
        L_{ID} = - y^g_i log(q^g_i)  + \frac{1}{l} \sum_{k=1}^l - y^{p_k}_i log(q^{p_k}_i).
    \end{aligned}
\end{equation}

\textbf{Overall object function.} The overall loss function of our methods is then:
\begin{equation}
    L=L_{SCC}+L_{WM}+L_{ID}.
\end{equation}
\section{Experiment}
\label{sec:experiment}

\subsection{Datasets and Evaluation Protocols}

We evaluate our proposed method on Market-1501 \cite{zheng2015scalable} and MSMT17 \cite{wei2018person}. Market-1501 includes 32,668 images of 1,501 identities with 6 camera views, 12,936 images of 751 identities for training, and 19,732 images of 750 identities for testing. MSMT17 is a more challenging dataset comprising 126,441 images of 4,101 person identities captured from 15 cameras. The training set has 32,621 images of 1,041 identities, and the test set has 93,820 images of 3,060 identities. We use mean average precision (mAP) and cumulative matching characteristic (CMC) Rank-1, Rank-5, Rank-10 accuracies as the evaluation metrics. There are no postprocessing operations in testing, such as reranking.

\subsection{Implementation Details}

All our experiments were conducted on four 2080Ti. The backbone of our Re-ID model is ResNet-50 \cite{he2016deep} pretrained on SYNTH-PEDES \cite{zuo2023plip}. We remove all the layers after the fourth bottleneck of the backbone network and set the stride of the first convolution layer of the fourth bottleneck to 1. We append a global average pooling layer and a 3x3 convolution layer as the part classifier after the backbone. We add a batch normalization layer and $l2$-normalization layer to obtain 2048-dimension features. The person images are resized to 384 $\times$ 128. Adam with weight decay of 5e-4 is adopted for training. The initial learning rate is set to 3.5e-4 and decreased by 10 after every 20 epochs. We train for a total 60 epochs and 400 iterations for each epoch. Following previous work \cite{ge2020self,cho2022part,wu2022multi,zhang2022implicit}, we employ DBSCAN based on Jaccard distance with k-reciprocal encoding for clustering. The search radius in DBSCAN is set to 0.5 in Market-1501 and 0.7 in MSMT17. The number of parts is set to 4 in Market-1501 and 7 in MSMT17. The semantic parsing label update factor $\gamma$ is set to 0.2 and decreases to 0 after 20 epochs. The memory update factor $m$ is set to 0.2. The temperature $\tau$ is set to its default value of 0.05. The part distillation parameter $\beta$ is set to 0.35.

\subsection{Ablation Study}

To validate the effectiveness of our method, we conduct detailed comparative experiments on Market-1501 and MSMT17. We adopt Part-based Pseudo Label Refinement \cite{cho2022part} with backbone changed to SYNTH-PEDES pretrained ResNet-50 as the baseline for our experiment. We verify the effectiveness of the proposed methods: spatial cascaded clustering (SCC) and weighted memory (WM). In conclusion, our method improves mAP of the baseline by 0.9\% and 3.0\% on Market-1501 and MSMT17, respectively.

\begin{table}[t]
\centering
\begin{tabular}{l|cc|cc}
\hline
\multirow{2}{*}{Methods} & \multicolumn{2}{c|}{Market-1501} & \multicolumn{2}{c}{MSMT17} \\ \cline{2-5} 
                         & mAP            & R1             & mAP          & R1          \\ \hline
baseline                 & 86.6           & 94.5           & 46.1         & 73.4        \\
$+L_{wNCE}$              & 87.2            & 94.4           & 47.2         & 74.6        \\
$+L_{WM}$                & 87.4            & 94.8           & 47.9         & 76.1           \\
$+L_{WM}+L_{parsing}$    & 84.4            & 93.5           & 41.4         & 70.0        \\
$+L_{WM}+L_{SCC}$         & \textbf{87.5}            & \textbf{94.9}           & \textbf{49.1}         & \textbf{76.5}        \\ \hline
\end{tabular}
\caption{Ablation study of each component in our proposed spatial cascaded clustering and weighted memory method.}
\label{tab:ablation}

\end{table}

\begin{figure}[t]
  \centering
  \begin{subfigure}[t]{0.2\linewidth}
    \centering
    \includegraphics[width=0.8\linewidth]{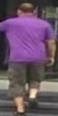}
    \caption{}
  \end{subfigure}\begin{subfigure}[t]{0.2\linewidth}
    \centering
    \includegraphics[width=0.8\linewidth]{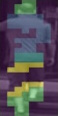}
    \caption{}
  \end{subfigure}\begin{subfigure}[t]{0.2\linewidth}
    \centering
    \includegraphics[width=0.8\linewidth]{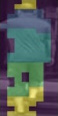}
    \caption{}
  \end{subfigure}\begin{subfigure}[t]{0.2\linewidth}
    \centering
    \includegraphics[width=0.8\linewidth]{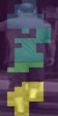}
    \caption{}
  \end{subfigure}\begin{subfigure}[t]{0.2\linewidth}
    \centering
    \includegraphics[width=0.8\linewidth]{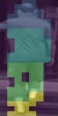}
    \caption{}
  \end{subfigure}
  \vspace{-3mm}
  \caption{Visualization of pseudo parsing masks: (a) Input image; (b) origin cascaded clustering; (c) origin with foreground correction; (d) origin with space correction; (e) our foreground and space corrected clustering.}
  \label{fig:clustercompare}

\end{figure}

\begin{figure}[t]
  \centering
  \begin{subfigure}[t]{0.48\linewidth}
    \centering
    \includegraphics[width=0.94\linewidth]{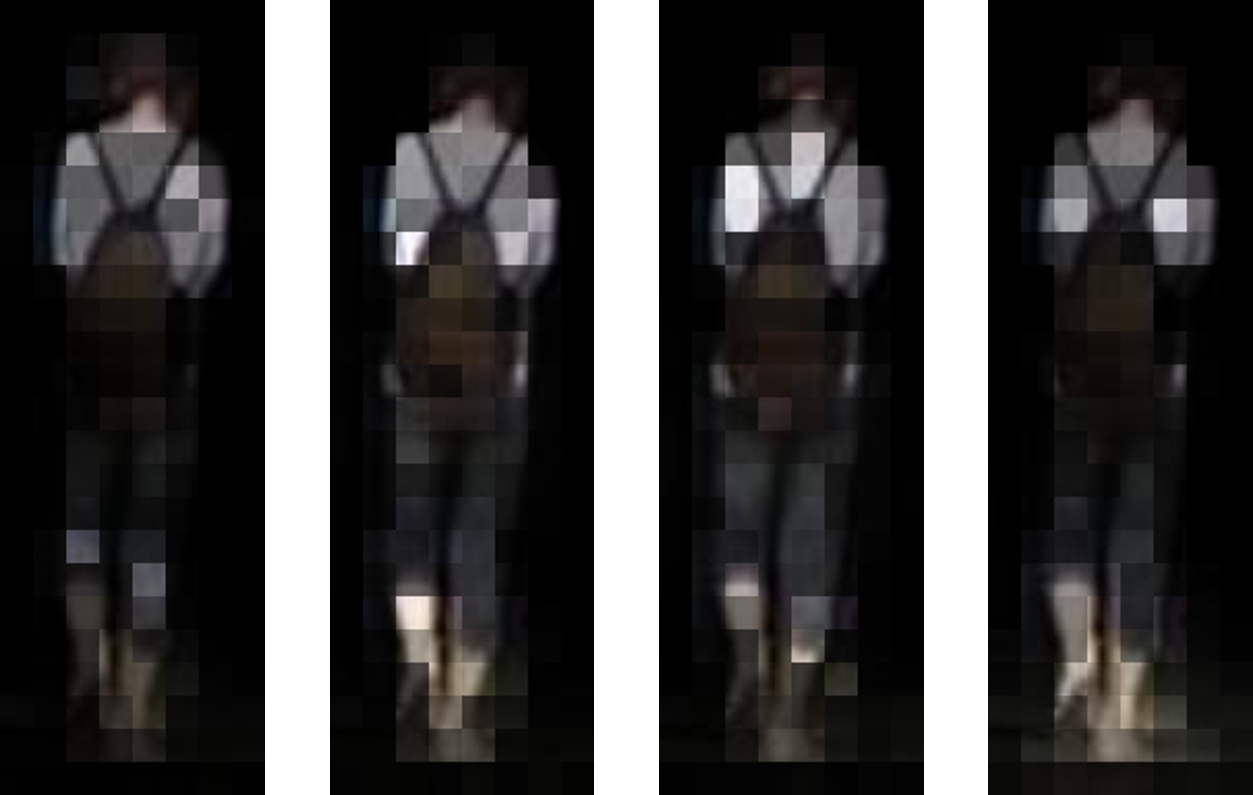}
    \caption*{Without $L_{diversity}$}
  \end{subfigure}\begin{subfigure}[t]{0.04\linewidth}
    \centering
    \includegraphics[width=0.1\linewidth]{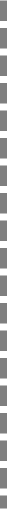}
  \end{subfigure}\begin{subfigure}[t]{0.48\linewidth}
    \centering
    \includegraphics[width=0.94\linewidth]{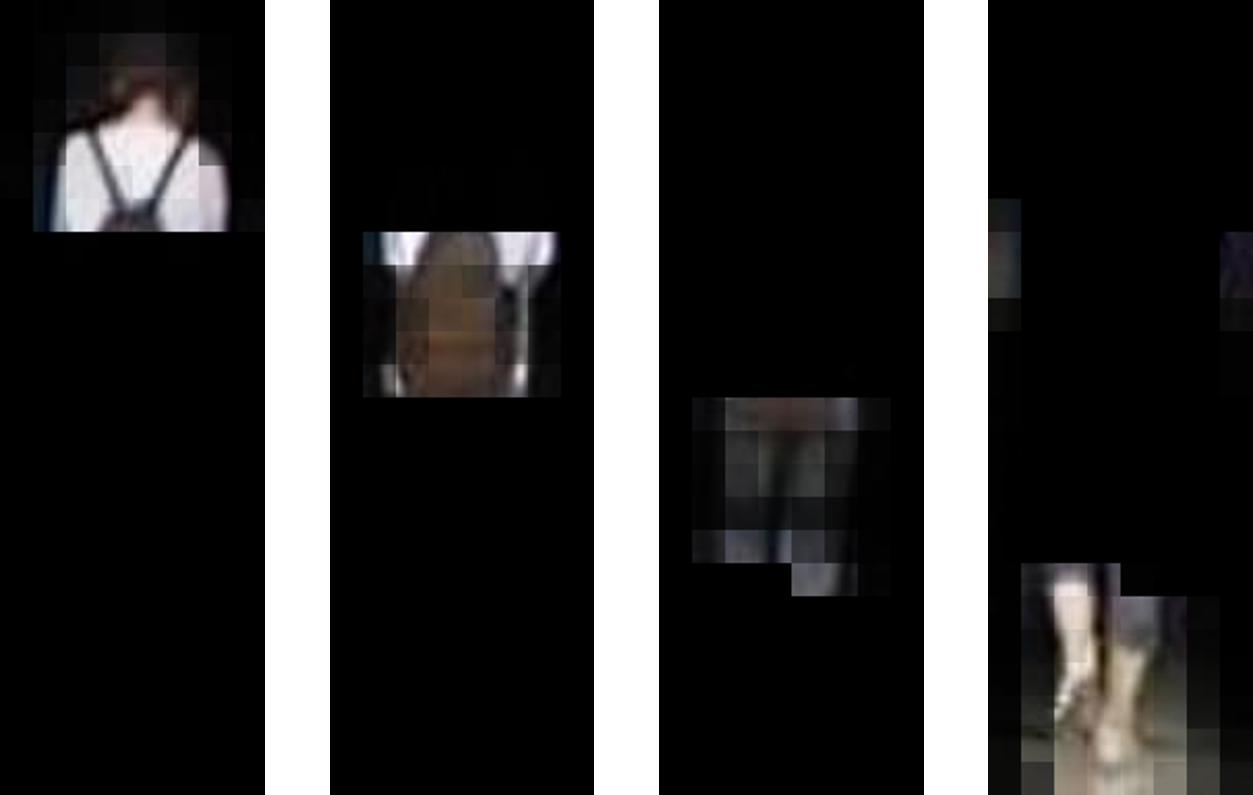}
    \caption*{With $L_{diversity}$}
  \end{subfigure}
  \vspace{-3mm}
  \caption{Visualization of predicted parsing masks. Without $L_{diversity}$, the predicted parsing masks the part classifier generates may degrade.}
  \label{fig:ldiversity}
  \vspace{-3mm}
\end{figure}

\textbf{Effectiveness of spatial cascaded clustering.} By comparing the results of \cref{tab:ablation} and \cref{tab:parsing}, we can observe the effectiveness of SCC. Thanks to our proposed foreground and space corrections, SCC can better extract part features of the person based on the similarity of feature vectors in feature space and spatial space. It improves the situation of semantic misalignment at the feature level. To demonstrate the effectiveness of our proposed improved cascade clustering, we visualize the parsing results given by the original, the foreground-corrected, the space-corrected, and the foreground and space-corrected cascade clustering methods, respectively. As shown in \cref{fig:clustercompare}, our two improvements allow cascade clustering to obtain more complete foreground information and parsing more consistent with spatial relationships. The origin parsing results often exhibit partial omissions and confusion between semantic parts. The possible reason is that the original cascaded clustering is applied to feature maps extracted by HRNet \cite{zhu2020identity}. Compared to the highly abstract feature maps of ResNet50, HRNet low-dimensional features are better suited for parsing tasks. So, when applied to ResNet50 feature maps, the original cascade clustering performs poorly. Another critical element in SCC is the $L_{diversity}$ loss. We also visualize the person parsing masks generated by the part classifier before and after using $L_{diversity}$ in \cref{fig:ldiversity}. As can be seen from the figure, without $L_{diversity}$, the parsing masks generated by the part classifier tend to be the same for each part, which significantly damages the diversity of part features. 

\textbf{Effectiveness of weighted memory.} By comparing the results of \cref{tab:ablation} and \cref{tab:memory}, we can observe the effectiveness of WM. The comparative experiment \cref{tab:memory} is conducted under completely identical and comprehensive conditions, with the only change being replacing the memory module update strategy with average and hardest \cite{dai2022cluster} updates. From \cref{tab:memory}, it can be seen that both the average update strategy and the hardest update strategy perform worse than our weighted update strategy. The average update strategy may be effective in most cases, but its effectiveness is limited due to its lack of focus on challenging sample exploration and noise suppression. The hardest update strategy provides an approach that focuses on exploring hard samples. However, the most hard samples will likely become the noise that pollutes the cluster centroid in the case of more noise samples. Our weighted update strategy lies between average updates and hardest updates. It neither treats all samples equally nor favors any one sample. Therefore, it can balance noise suppression and hard sample mining. 

\begin{table}[t]
\centering
\begin{tabular}{l|cc|cc}
\hline
\multirow{2}{*}{Methods} & \multicolumn{2}{c|}{Market-1501} & \multicolumn{2}{c}{MSMT17} \\ \cline{2-5} 
                         & mAP            & R1             & mAP          & R1          \\ \hline
cascaded clustering      & 86.2           & 94.4           & 48.8            & 75.9           \\
foreground correction    & 86.9           & \textbf{95.1}           & 48.7         & 76.0        \\
space correction         & 86.5           & 94.4           & \textbf{49.3}         & \textbf{77.0}        \\
both correction          & \textbf{87.5}           & 94.9           & 49.1         & 76.5        \\ \hline
\end{tabular}
\caption{Comparison of different parsing methods for part feature extraction.}
\label{tab:parsing}
\vspace{-3mm}
\end{table}

\begin{figure}[t]
  \centering\begin{subfigure}[t]{0.5\linewidth}
    \centering
    \includegraphics[width=\linewidth]{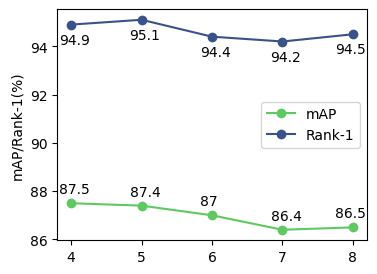}
    \caption{$l$}
  \end{subfigure}\begin{subfigure}[t]{0.5\linewidth}
    \centering
    \includegraphics[width=\linewidth]{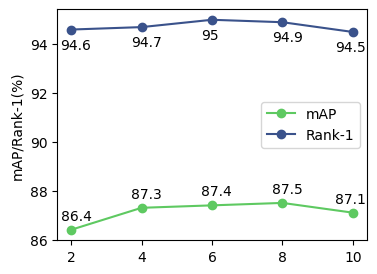}
    \caption{$\eta$}
  \end{subfigure}
  \caption{Parameter analysis of $\eta$ and $l$ on Market-1501.}
  \label{fig:para}
  \vspace{-3mm}
\end{figure}

\textbf{Parameter Analyse.} We analyze two parameters in our method: the distance threshold $\eta$ and the part numbers $l$ in spatial cascaded clustering. We keep one parameter constant and adjust the other parameter for experimentation, and the results are in \cref{fig:para}. A smaller $\eta$ value will make the clustering conditions more stringent. When $\eta$ is too small, clustering may absorb some vectors with a large spatial distance, causing the spatial constraints to become ineffective. Similarly, when $\eta$ is too large, it is equivalent to having no spatial constraints. As shown in \cref{fig:para}, it can be observed that when $\eta$ is relatively large or small, there is a slight decrease in the mAP metric. Another critical parameter is part number $l$. On Market-1501, the best performance is achieved when $l=4$. That is, the foreground is divided into three parts. This result is similar to previous work \cite{zhu2020identity}. The probable reason is that in the Market-1501 dataset, people’s clothing is relatively unified, mainly divided into tops, bottoms, and shoes, and the clothing colors are vibrant. Therefore, additional personal items such as backpacks, handbags, bicycles, etc., may not significantly influence the classification. Howerver, on MSMT17, the best performance is achieved when $l=7$. This may be due to the prevalence of occlusions and incomplete cropping in person images within this dataset. 

\begin{table}[t]
\centering
\begin{tabular}{l|cc|cc}
\hline
\multirow{2}{*}{Methods} & \multicolumn{2}{c|}{Market-1501} & \multicolumn{2}{c}{MSMT17} \\ \cline{2-5} 
                         & mAP            & R1             & mAP          & R1          \\ \hline
average update           & 87.2           & 94.7           & 48.7         & 76.0        \\
hardest update           & 87.4           & \textbf{95.1}           & 48.3         & 75.6        \\
weighted update          & \textbf{87.5}           & 94.9           & \textbf{49.1}         & \textbf{76.5}        \\ \hline
\end{tabular}
\caption{Comparison of different update strategies for memory module.}
\label{tab:memory}
\vspace{-3mm}
\end{table}

\begin{figure}[t]
  \centering
  \includegraphics[width=\linewidth]{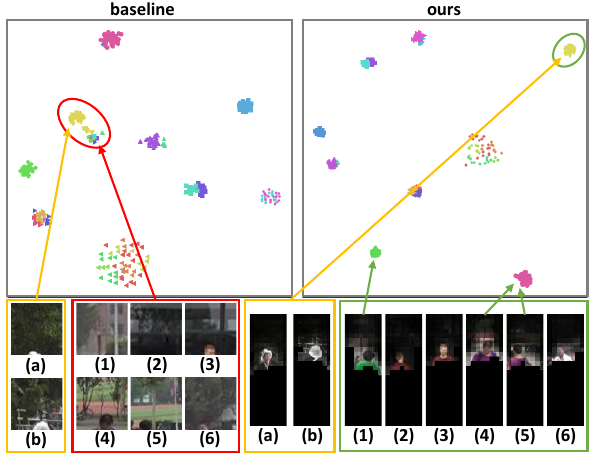}
  \caption{The t-SNE \cite{van2008tsne} visualization of baseline and ours clustering results on same samples. Samples with the same color in both graphs have the same ground true label. The uppermost portion split by the baseline's horizontal partitioning is shown on the left, while the top portion masks extracted by our method are shown on the right. Typical samples (a) and (b) have the same ID, while (1) - (6) are from other IDs. The red and green circled regions show that samples mixed in the baseline are separated in our method.}
  \label{fig:tsne}
  \vspace{-3mm}
\end{figure}

\begin{table*}[t]
\centering
\begin{tabular}{lccccccccc}
\hline
\multicolumn{1}{l|}{\multirow{2}{*}{Method}} & \multicolumn{1}{c|}{\multirow{2}{*}{Reference}} & \multicolumn{4}{c|}{Market-1501}                & \multicolumn{4}{c}{MSMT17} \\ \cline{3-10} 
\multicolumn{1}{l|}{}                        & \multicolumn{1}{c|}{}                           & mAP  & R1   & R5   & \multicolumn{1}{c|}{R10}  & mAP   & R1   & R5   & R10  \\ \hline
\textit{Purely Unsupervised}                           &                                                 &      &      &      &                           &       &      &      &      \\ \hline
\multicolumn{1}{l|}{BUC \cite{lin2019bottom}}                     & \multicolumn{1}{c|}{AAAI'19}                    & 38.3 & 66.2 & 79.6 & \multicolumn{1}{c|}{84.5} & -     & -    & -    & -    \\
\multicolumn{1}{l|}{MMCL \cite{wang2020unsupervised}}                    & \multicolumn{1}{c|}{CVPR'20}                    & 45.5 & 80.3 & 89.4 & \multicolumn{1}{c|}{92.3} & 11.2  & 35.4 & 44.8 & 49.8 \\
\multicolumn{1}{l|}{HCT \cite{zeng2020hierarchical}}                     & \multicolumn{1}{c|}{CVPR'20}                    & 56.4 & 80.0 & 91.6 & \multicolumn{1}{c|}{95.2} & -     & -    & -    & -    \\
\multicolumn{1}{l|}{MMT \cite{ge2020mutual}}                     & \multicolumn{1}{c|}{ICLR'20}                    & 74.3 & 88.1 & 96.0 & \multicolumn{1}{c|}{97.5} & -     & -    & -    & -    \\
\multicolumn{1}{l|}{SpCL \cite{ge2020self}}                    & \multicolumn{1}{c|}{NeurIPS'20}                 & 73.1 & 88.1 & 95.1 & \multicolumn{1}{c|}{97.0} & 19.1  & 42.3 & 56.5 & 68.4 \\
\multicolumn{1}{l|}{GCL \cite{chen2021joint}}                     & \multicolumn{1}{c|}{CVPR'21}                    & 66.8 & 87.3 & 93.5 & \multicolumn{1}{c|}{95.5} & 21.3  & 45.7 & 58.6 & 64.5 \\
\multicolumn{1}{l|}{ICE \cite{chen2021ice}}                     & \multicolumn{1}{c|}{ICCV'21}                    & 79.5 & 92.0 & 97.0 & \multicolumn{1}{c|}{98.1} & 29.8  & 59.0 & 71.7 & 77.0 \\
\multicolumn{1}{l|}{RLCC \cite{zhang2021refining}}                    & \multicolumn{1}{c|}{CVPR'21}                    & 77.7 & 90.8 & 96.3 & \multicolumn{1}{c|}{97.5} & 27.9  & 56.5 & 68.4 & 73.1 \\
\multicolumn{1}{l|}{MCRN \cite{wu2022multi}}                    & \multicolumn{1}{c|}{AAAI'22}                    & 80.8 & 92.5 & -    & \multicolumn{1}{c|}{-}    & 31.2  & 63.6 & -    & -    \\
\multicolumn{1}{l|}{CCL \cite{dai2022cluster}}                     & \multicolumn{1}{c|}{ACCV'22}                    & 82.1 & 92.3 & 96.7 & \multicolumn{1}{c|}{97.9} & 27.6  & 56.0 & 66.8 & 71.5 \\
\multicolumn{1}{l|}{PPLR \cite{cho2022part}}                    & \multicolumn{1}{c|}{CVPR'22}                    & 81.5 & 92.8 & 97.1 & \multicolumn{1}{c|}{98.1} & 31.4  & 61.1 & 73.4 & 77.8 \\
\multicolumn{1}{l|}{ISE \cite{zhang2022implicit}}                     & \multicolumn{1}{c|}{CVPR'22}                    & 84.7 & 94.0 & 97.8 & \multicolumn{1}{c|}{98.8} & 35.0  & 64.7 & 75.5 & 79.4 \\ \hline
\multicolumn{1}{l|}{ISE*}                    & \multicolumn{1}{c|}{-}                          & \textbf{87.6} & \textbf{95.3} & \underline{98.2} & \multicolumn{1}{c|}{\underline{98.9}} & \underline{46.4}  & \underline{74.9} & \underline{84.3} & \underline{87.5} \\
\multicolumn{1}{l|}{PPLR* (Baseline)}         & \multicolumn{1}{c|}{-}                          & 86.6 & 94.5 & 98.0 & \multicolumn{1}{c|}{98.7} & 46.1  & 73.4 & 83.7 & 87.0 \\
\multicolumn{1}{l|}{\textbf{Our SCWM*}}              & \multicolumn{1}{c|}{This paper}                          & \underline{87.5} & \underline{94.9} & \textbf{98.4} & \multicolumn{1}{c|}{\textbf{99.0}} & \textbf{49.1}  & \textbf{76.5} & \textbf{85.4} & \textbf{88.4} \\ \hline
\textit{Supervised}                                   &                                                 &      &      &      &                           &       &      &      &      \\ \hline
\multicolumn{1}{l|}{PCB \cite{sun2018beyond}}                     & \multicolumn{1}{c|}{ECCV'18}                    & 81.6 & 93.8 & 97.5 & \multicolumn{1}{c|}{98.5} & 40.4  & 68.2 & -    & -    \\
\multicolumn{1}{l|}{ISP \cite{zhu2020identity}}                     & \multicolumn{1}{c|}{ECCV'20}                    & 88.6 & 95.3 & 98.6 & \multicolumn{1}{c|}{-}    & -     & -    & -    & -    \\ 
\multicolumn{1}{l|}{ABD-Net \cite{chen2019abd}}                     & \multicolumn{1}{c|}{ICCV'19}                    & 88.3 & 95.6 & - & \multicolumn{1}{c|}{-}    & 60.8     & 82.3    & -    & -    \\
\multicolumn{1}{l|}{ISE (w/ ground-truth) \cite{zhang2022implicit}}                     & \multicolumn{1}{c|}{CVPR'22}                    & 87.8 & 95.6 & 98.5 & \multicolumn{1}{c|}{99.2}    & 51.0     & 76.8    & 87.1    & 90.6    \\ 
\multicolumn{1}{l|}{\textbf{Our SCWM* (w/ ground-truth)}}                     & \multicolumn{1}{c|}{This paper}           & 88.1 & 95.3 & 98.3 & \multicolumn{1}{c|}{99} & 60.4  & 82.3 & 90.4  & 92.7    \\ \hline
\end{tabular}
\caption{Comparison with state-of-the-art re-ID methods on Market-1501 and MSMT17 datasets. The first and second best results among all unsupervised methods are marked in \textbf{bold} and \underline{underlined}, respectively. * denotes the backbone settings with SYNTH-PEDES pretrained ResNet-50.}
\label{tab:sota}
\vspace{-3mm}
\end{table*}

\textbf{Clustering Quality.} We visualize the baseline's and our clustering results in \cref{fig:tsne}. Due to the different poses of the person, the top horizontal partitioning used by the baseline to extract part features mainly contains the background. This results in a person with different labels incorrectly clustered due to the background similarity. In contrast, our method focuses on discriminative head features of the person with the masks obtained through self-learning parsing, making distinguishing persons with different identities more precise. From \cref{fig:tsne}, it can be observed that for those samples mixed into other clusters, we can generally find their actual categories using our method and maintain the original clustering more closely and cleanly.

\subsection{Comparison with State-of-the-Arts}

We compare the proposed SCWM with various state-of-the-art unsupervised person re-ID methods on two standard datasets, e.g. Market-1501 and MSMT17. All results are shown in \cref{tab:sota}. For fairness, we also test the performance of the PPLR \cite{cho2022part} and ISE \cite{zhang2022implicit} methods on our backbone separately, and the results are marked with asterisks in the table. From the table, it can be seen that our proposed SCWM method surpasses most previous unsupervised methods, including BUC \cite{lin2019bottom}, MMCL \cite{wang2020unsupervised}, HCT \cite{zeng2020hierarchical}, MMT \cite{ge2020mutual}, GCL \cite{chen2021joint}, ICE \cite{chen2021ice}, SpCL \cite{ge2020self}, RLCC \cite{zhang2021refining}, MCRN \cite{wu2022multi}, CCL \cite{dai2022cluster}, and PPLR \cite{cho2022part}. In addition, compared to supervised part-based methods, we have also achieved similar performance. Only PPLR, PCB \cite{sun2018beyond}, ISP \cite{zhu2020identity}, and our SCWM in the table are part-based methods, and others only focus on global features. Among these part-based methods, only we individually apply metric learning to each part feature. PPLR and PCB do not utilize metric learning, while ISP concatenates all part features for metric learning. ISE \cite{zhang2022implicit} proposes a method for sample augmentation in the feature space. However, our backbone already provides good clustering features, which narrows the gap between ISE and other methods. Therefore, our method surpasses them by being able to extract and leverage more abundant local information. As shown in \cref{tab:sota}, our method surpasses the baseline method on all the benchmarks with +0.9\% and 3.0\% of mAP on Market-1501 and MSMT17, respectively. On MSMT17, our method significantly outperforms the prior state-of-the-art method \cite{zhang2022implicit} with 2.7\% higher in mAP. In addition, we also test the results of our method with ground true labels. As shown in the lower block of \cref{tab:sota}, our method is competitive with previous supervised part-based methods and surpass ISE \cite{zhang2022implicit} with ground truth. Our method achieves a similar performance to ABD-Net in the supervised setting. These results indicate that our method is also applicable to supervised re-ID tasks.

\section{Conclusion}

In part-based unsupervised person re-identification methods, the primary challenges stem from misalignment and noise of part features. To address these two issues, we introduced self-learning human parsing with spatial cascaded clustering, which allows part features to be semantically aligned and improves the precision of part feature segmentation to reduce noise. We analyzed two limitations of the cascaded clustering method used in current approaches: foreground omissions and spatial confusions. We proposed foreground and space correction, significantly improving the parsing performance. After aligning part features, we revised the memory module based on the proposed assumptions and requirements to better learn and leverage part features. We propose a weighted memory with a flexible weighted update strategy considering hard example mining and noise suppression. Experiments’ results show that our method has achieved almost the best performance compared to state-of-the-art methods.
{
    \small
    \bibliographystyle{ieeenat_fullname}
    \bibliography{main}
}


\end{document}